\documentclass[final]{cvpr}

\usepackage{times}
\usepackage{epsfig}
\usepackage{graphicx}
\usepackage{amsmath}
\usepackage{amssymb}
\usepackage{comment}

\usepackage{hhline}
\usepackage{multirow}
\usepackage{amsfonts}
\usepackage{bbm}
\usepackage[ruled]{algorithm2e}
\usepackage{graphicx}

\usepackage{floatrow}
\newfloatcommand{capbtabbox}{table}[][\FBwidth]

\usepackage[pagebackref=true,breaklinks=true,colorlinks,bookmarks=false]{hyperref}
\let\footnote\thanks
\def\cvprPaperID{2534} % *** Enter the CVPR Paper ID here
\def\confYear{CVPR 2021}

\begin{document}

%%%%%%%%% TITLE
\title{FedDG: Federated Domain Generalization on Medical Image Segmentation \\via  Episodic Learning in Continuous Frequency Space}

%\author{Paper ID: 2534}

\author{Quande Liu$^1$, Cheng Chen$^1$, Jing Qin$^2$, Qi Dou$^{1,}$\footnote{Corresponding author}~, Pheng-Ann Heng$^1$\\
$^1$~Department of Computer Science and Engineering, The Chinese University of Hong Kong\\
$^2$~School of Nursing, The Hong Kong Polytechnic University\\
{\tt\small \{qdliu, cchen, qdou, pheng\}@cse.cuhk.edu.hk, harry.qin@polyu.edu.hk}
% \and
}

\maketitle

%%%%%%%%% ABSTRACT
\begin{abstract}

Federated learning allows distributed medical institutions to collaboratively learn a shared prediction model with privacy protection. 
While at clinical deployment, the models trained in federated learning can still suffer from performance drop when applied to completely unseen hospitals outside the federation.
In this paper, we point out and solve a novel problem setting of federated domain generalization (FedDG), which aims to learn a federated model from multiple distributed source domains such that it can directly generalize to unseen target domains. 
We present a novel approach, named as Episodic Learning in Continuous Frequency Space (ELCFS), for this problem by enabling each client to exploit multi-source data distributions under the challenging constraint of data decentralization.
Our approach transmits the distribution information across clients in a privacy-protecting way through an effective continuous frequency space interpolation mechanism.
With the transferred multi-source distributions, we further carefully design a boundary-oriented episodic learning paradigm to expose the local learning to domain distribution shifts and particularly  meet the  challenges of model generalization in medical image segmentation scenario.
The effectiveness of our method is demonstrated with superior performance over state-of-the-arts and in-depth ablation experiments on two medical image segmentation tasks. The code is available at~\href{https://github.com/liuquande/FedDG-ELCFS}{https://github.com/liuquande/FedDG-ELCFS}.

\end{abstract}

%%%%%%%%% BODY TEXT
% \vspace{-3mm}
\section{Introduction}
% Data-driven machine learning (ML) has emerged as a promising approach for building accurate and robust statistical models from medical data, which is collected in huge volumes by modern healthcare systems. Existing medical data is not fully exploited by ML primarily because it sits in data silos and privacy concerns restrict access to this data. However, without access to sufficient data, ML will be prevented from reaching its full potential and, ultimately, from making the transition from research to clinical practice. This paper considers key factors contributing to this issue, explores how federated learning (FL) may provide a solution for the future of digital health and highlights the challenges and considerations that need to be addressed.

% Data generated by networks of mobile and IoT devices poses unique challenges for training machine
% learning models. Due to the growing storage/computational power of these devices and concerns about
% data privacy, it is increasingly attractive to keep data and computation locally on the device (Smith
% et al., 2017). Federated Learning (FL) (Mohassel & Rindal, 2018; Bonawitz et al., 2017; Mohassel
% & Zhang, 2017) provides a privacy-preserving mechanism to leverage such decen-tralized data and
% computation resources to train machine learning models. The main idea behind federated learning is
% to have each node learn on its own local data and not share either the data or the model parameters.

Data collaboration across multiple medical institutions is increasingly desired to build accurate and robust data-driven deep networks for medical image segmentation~\cite{dhruva2020aggregating,kaissis2020secure,shilo2020axes}.
Federated learning (FL)~\cite{konevcny2016federated} has recently opened the door for a promising privacy-preserving solution, which allows training a model on distributed datasets while keeping data locally.
The paradigm works in a way that each local client (e.g., hospital) learns from their own data, and only aggregates the model parameters at a certain frequency at the central server to generate a global model. 
% The general principle is to let each local client (i.e. hospital) learn from their own data, and only aggregate the model parameters at a certain frequency to generate a global model shared by all clients. 
All the data samples are kept within each local client during federated training.

\begin{figure}[t]
	\centering
	\includegraphics[width=0.94\textwidth]{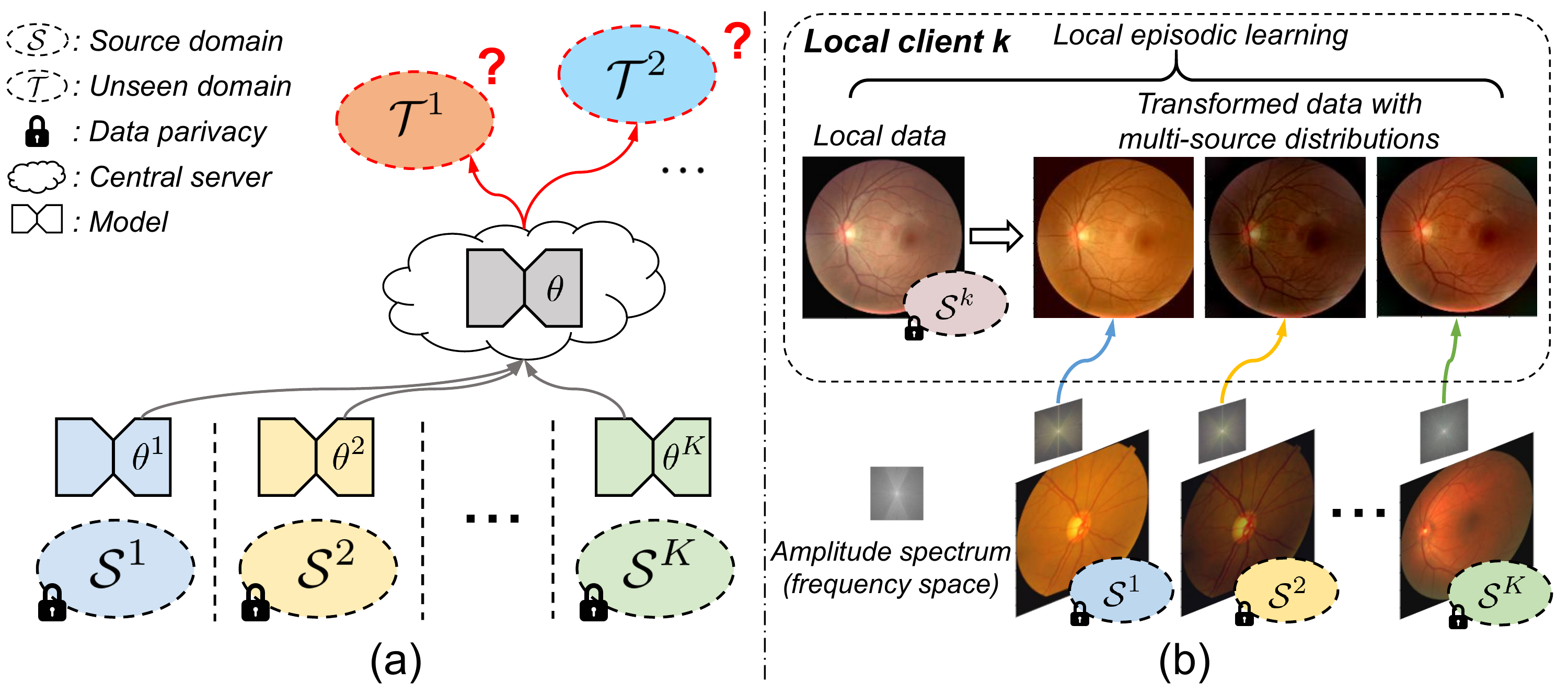}
% 	\caption{(a) Illustration of the problem of federated domain generalization, which aims to learn a federated model from multiple distributed source domains such that it can directly generalize to completely unseen domains. (b) Our solution to locally simulate the multi-source data distributions. By transferring the frequency space signals across clients, we transform the appearance of local data ($x^k$) to diverse distributions (i.e., style) of other clients, producing $x^{k\rightarrow 1}, x^{k\rightarrow 2},\cdots, x^{k\rightarrow K} $. The simulated local distributions enables each client to take the potential of joint statistics information to learn towards domain-invariance, hence can improve the generalizability of federated model. (Best zoom in and view in color).} 
	\caption{(a) The novel problem setting of federated domain generalization (FedDG), which aims to learn a federated model from multiple decentralized source domains such that it can directly generalize to completely unseen target domains. (b) Our main idea to tackle FedDG by transferring distribution information in  frequency space and episodic learning at each local client.
	%Through transferring the distribution information detached from frequency space signals across clients, the local data can be transformed to various data distributions (i.e., style), allowing the local learning to exploit joint distribution information and to improve generalizability. (Best zoom in and view in color.)
	} 
	\label{fig:intro}
	\vspace{-2mm}
\end{figure}

Although FL has witnessed some pilot progress on medical image segmentation tasks~\cite{chang2020synthetic,rieke2020future,sheller2018multi}, 
% all existing works only focus on improving model's performance on internal clients.
% While model's generalizability on completely unseen hospitals 
% neglecting model generalizability on completely unseen hospitals outside the federation. Particularly, in clinical model deployment, the medical images encountered in unseen hospitals can differ significantly with the source clients in their data distributions, due to the difference in imaging scanners and protocols. 
all existing works only focus on improving model performance on the internal clients, while neglecting model generalizability onto unseen domains outside the federation. 
This is a crucial problem impeding wide applicability of FL models in real practice.
The testing medical images encountered in unseen hospitals can differ significantly from the source clients in terms of data distributions, due to the variations in imaging scanners and protocols. 
%In clinical model deployment, 
How to generalize the federated model under such distribution shifts is technically challenging yet unexplored so far.
% While how to generalize the federated model to such distribution shift is less investigated yet unsolved.
In this work, we identify the novel problem setting of ~\textit{Federated Domain Generalization} (FedDG), which aims to learn a federated model from multiple decentralized source domains such that it can directly generalize to completely unseen domains, as illustrated in Fig.~\ref{fig:intro} (a).

Unseen domain generalization (DG) is an active research topic with different methods being proposed~\cite{carlucci2019domain,dou2019domain,ghifary2015domain,li2017learning,li2019episodic,li2018domain,li2018deep,motiian2017unified,qiao2020learning}, but the federated paradigm with distributed data sources poses new challenges for DG. With the goal to extract representations that are robust to distribution shift, existing DG approaches usually require access to multi-source distributions in the learning process. 
% For example, adversarial feature alignment methods~\cite{li2018domain,li2018deep} have to obtain sample features from different distributions to promote the learning of domain-invariant representations. 
% Meta-learning based methods~\cite{dou2019domain,li2017learning} need to use multi-source data simultaneously, which are split into meta-train and meta-test to expose the training to domain shift.
For instance, adversarial feature alignment methods~\cite{li2018domain,li2018deep} have to train the domain discriminator with samples from different source datasets. 
Meta-learning based methods~\cite{dou2019domain,li2017learning} need to use multi-source data of different distributions to construct virtual training and virtual testing domains within each minibatch.
% Meta-learning based methods~\cite{dou2019domain,li2017learning} need to use multi-source data with different distributions in each training iteration to expose the optimization process to domain shift.
% Access to multi-domain data is required in meta-learning-based methods~\cite{li2017learning,dou2019domain}, which split them into meta-train and meta-test at each iteration to simulate the domain shift. 
Whereas in federated paradigm, data are stored distributedly and \textit{the learning at each client can only access its local data}.
% which constrains the local learning to make full use of the multi-source distributions to obtain domain-invariance.
Therefore, current DG methods are typically not applicable in FedDG scenario. In addition, the local optimization would make model biased to its own data distribution, thus less generalizable to new target domains.

To solve this FedDG problem, our insight is to enable each client to access multi-source data distributions in a privacy-protecting way. 
The idea is motivated by the knowledge that the low-level distributions (i.e., style)  and high-level semantics of an image can be respectively captured by amplitude and phase spectrum in the frequency space, as revealed by visual psychophysics ~\cite{guyader2004image,piotrowski1982demonstration,yang2020phase}.
We can consider exchanging these amplitude spectrum across clients to transmit the distribution information (cf. Fig.~\ref{fig:intro} (b)), while keeping the phase spectrum with core semantics locally for privacy protection.
Based on this, we also devise a continuous frequency space interpolation mechanism, which interpolates between the local and transferred distributions for enriching the established multi-domain distributions for each local client.
This promotes the local training to gain domain-invariance benefiting from a dedicated dense distribution space.
With these established distributions, we expose the local learning to domain distribution shifts via an episodic training paradigm to enhance the generalizability of local parameters. A novel meta-update objective function is designed to guide cross-domain optimization attending to the boundary area. This is notably important for medical image segmentation applications where generalization errors often come from imprecise predictions at ambiguous boundary of anatomies.

Our main contributions are highlighted as follows:
\vspace{-1mm}
\begin{itemize}
    \item We tackle the novel and practical problem of \textit{Federated Domain Generalization}. To the best of our knowledge, this is the first work to improve generalizability on completely unseen domains for federated models.

\vspace{-1mm}

    %\item We propose a privacy-preserving way to extract the distribution information from different clients, and to to learn generalizable federated models from decentralized datasets

	\item We propose a privacy-preserving solution to learn the generalizable FL model under decentralized datasets, through an effective continuous frequency space interpolation mechanism across clients.
	
	\vspace{-1mm}
	
	\item We present a novel boundary-oriented episodic learning scheme for the local training at a client, which exposes local optimization to domain shifts and enhances model generalizability at ambiguous boundary area.
	
	\vspace{-1mm}
	
	\item We conduct extensive experiments on two typical medical image segmentation tasks, i.e., retinal fundus image segmentation (four datasets) and prostate MRI segmentation (six datasets). Our achieved superior performance over state-of-the-arts and in-depth analytical experiments demonstrate the efficacy of our approach.
	%Code will be released publicly.
\end{itemize}

\section{Related Work}
\subsection{Federated Learning in Medical Imaging}
% \subsection{Federated Learning in Healthcare}
% Federated learning (FL)~\cite{konevcny2016federated,mcmahan2017communication,yang2019federated} provides a privacy-preserving solution for data collaboration, in which a machine learning model is built based on datasets that are distributed across multiple local sites without data centralization.
Federated learning~\cite{hsu2020federated,konevcny2016federated,mcmahan2017communication,yang2019federated} provides a promising privacy-preserving solution for multi-site data collaboration, which develops a global model from decentralized datasets by aggregating the parameters of each local client while keeping data locally.
% Kone\v{c}n\'y et al.~\cite{konevcny2016federated} propose to learn a local update from a restricted space or to compress the learned updates before sending it to the server to improve the communication efficiency of FL.
Representatively, McMahan et al.~\cite{mcmahan2017communication} propose the popular federated averaging algorithm for communication-efficient federated training of deep networks.
% Representatively, McMahan et al.~\cite{mcmahan2017communication} propose the popular FederatedAveraging (FedAvg) algorithm which increases the computation of local updates between each communication round to improve the communication efficiency.
% FL has recently drawn increasing interests in medical image applications~\cite{chang2020synthetic,kaissis2020secure,lifederated,li2019privacy}, as FL helps alleviate the ethics and regulations concerns.
With the advantage of privacy protection, FL has recently drawn increasing interests in medical image applications~\cite{chang2020synthetic,kaissis2020secure,li2020federated,li2019privacy,roth2020federated,sheller2018multi,silva2019federated}.
% in which data sharing is highly sensitive and raises serious concerns.
Sheller et al.~\cite{sheller2018multi} is a pilot study to investigate the collaborative model training without sharing patient data for the multi-site brain tumor segmentation.
Later on, Li et al.~\cite{li2019privacy} further compare several weights sharing strategies in FL to alleviate the effect of data imbalance among different hospitals. 
However, these works all focus on improving performance on internal clients, without considering the generalization issue for unseen domains outside the federation, which is crucial for wide clinical usability.
Latest literature has studied a related problem of unsupervised domain adaptation in FL paradigm~\cite{li2020multi,peng2019federated}, whereas these methods typically require data from the target domain to adapt the model. In practice, it would be time-consuming or even impractical to collect data from each new hospital before model deployment.
Instead, our tackled new problem setting of FedDG aims to directly generalize the federated model to completely unseen domains, in which no prior knowledge from the target domain is needed.
% To our best knowledge, the work proposed here is the first federated learning framework to improve model generalizability on completely unseen domains.
 %Peng et al.~\cite{peng2019federated} extend adversarial domain adaptation to the setting of FL and leverage a dynamic attention mechanism and feature disentanglement to enhance knowledge transfer.
% Li et al.~\cite{li2020multi} study domain adaptation in federated multi-site fMRI analysis using the mixture of experts and adversarial domain alignment.

\subsection{Domain Generalization}
Domain generalization~\cite{chattopadhyay2020learning,du2020learning,gong2019dlow,hoffer2020augment,qiao2020learning,seo2019learning,yue2019domain,zakharov2019deceptionnet} aims to learn a model from multiple source domains such that it can directly generalize to unseen target domains. 
% There have been a variety of existing approaches attempting to learn generalizable models.
Among previous efforts, some methods aim to learn domain-invariant representations by minimizing the domain discrepancy across multiple source domains~\cite{ghifary2015domain,hsu2017learning,li2018domain,li2018deep,liu2020shape,motiian2017unified,muandet2013domain,wang2020dofe}.
% For example, Li et al.~\cite{li2018deep} propose a conditional invariant adversarial network that learns to minimize the discrepancy in the conditional distributions of labels given inputs across domains.
For example, Motiian et al.~\cite{motiian2017unified} utilize a contrastive loss to minimize the distance between samples from the same class but different domains.
% Li et al.~\cite{li2018domain} imposes the Maximum Mean Discrepancy measure into adversarial autoencoders to align the distributions of multiple domains.
Some other DG methods are based on meta-learning, which is an episodic training paradigm by creating meta-train and meta-test splits at each iteration to stimulate domain shift~\cite{balaji2018metareg,dou2019domain,li2017learning,li2019feature}.
Li et al.~\cite{li2019feature} employ meta-learning to learn an auxiliary loss that guides the feature extractor to learn more generalized features.
% Dou et al.~\cite{dou2019domain} further enforce semantic feature guidance in the episodic learning scheme by using a global class alignment and local sample clustering.
However, these methods typically require centralizing multi-domain data in one place for learning, which violates privacy protection in federated learning setting with decentralized datasets.

There are other methods tackling DG by manipulating deep neural network architectures~\cite{khosla2012undoing,li2017deeper,matsuura2020domain}, leveraging self-supervision signals~\cite{carlucci2019domain,wang2020learning}, designing training heuristics~\cite{huang2020self,li2019episodic}, or conducting data augmentations~\cite{shankar2018generalizing,volpi2018generalizing,zhang2020generalizing,zhou2020learning}, which are free from requirement of data centralization. Representatively, Carlucci et al.~\cite{carlucci2019domain} adopt self-supervised learning by solving jigsaw puzzles. Zhang et al.~\cite{zhang2020generalizing} conduct extensive data augmentations on each source domain by stacking a series of transformations.
These approaches, when applied in FL paradigm, can helpfully act as regularizations for the local training with individual source domain data, yet hardly exploit the rich data distributions across domains.
Our method instead, aims to transfer the distribution information across clients to make full use of the multi-source distributions towards FedDG.
We also experimentally compare with these typical methods under the FL setting with superior performance demonstrated.

% Our method instead, aim to explore the multiple decentralized source distributions and simulate diverse distributions locally towards federated domain generalization.
% We compare with \cite{carlucci2019domain} and \cite{zhang2020generalizing} under the FL setting and the experimental results demonstrate the superior performance of our method.

% \begin{figure*}[t]
% 	\centering
% 	\includegraphics[width=\textwidth]{figure/method.pdf}
% 	\caption{Our solution for federated domain generalization. In local learning at federated client, the local data is online transformed to multi-source distributions in a continuous space by interpolating the frequency space signal of amplitude spectrum with other clients. The raw local data and transformed data then serves as meta-train and meta-test to simulate real-world distribution shift, and the feature cohesion and separation are regularized among the boundary and background prototypes extracted from these data to encourage the learning towards domain-invariance. } 
% 	\label{fig:method}
% \end{figure*}
\begin{figure*}[t]
	\centering
	\includegraphics[width=\textwidth]{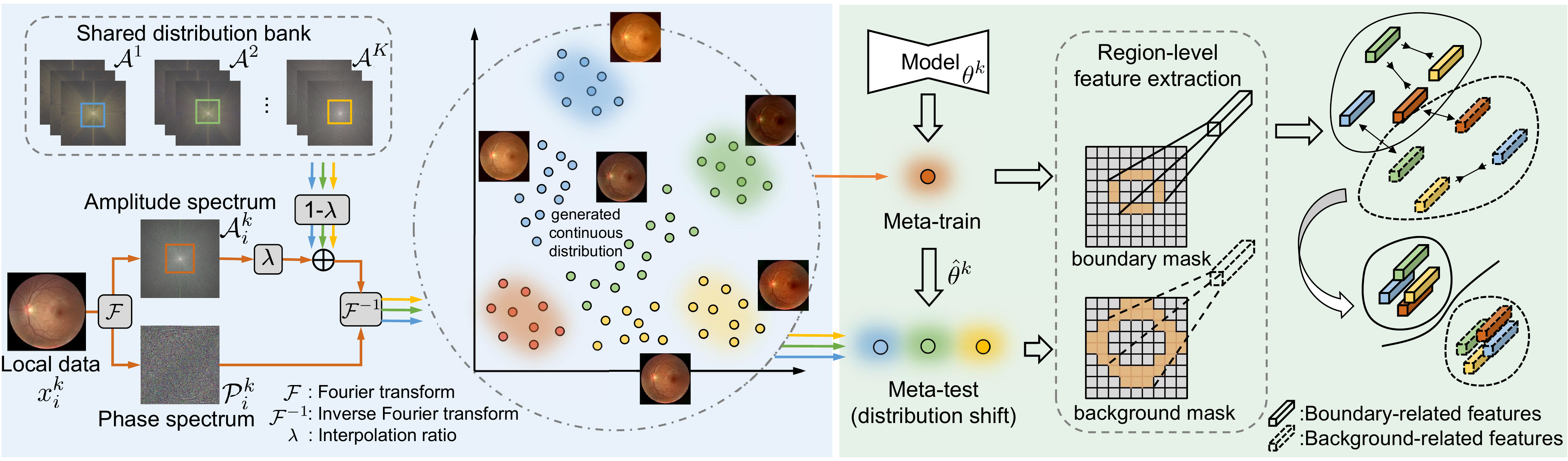}
	\caption{Overview of our proposed episodic learning in continuous frequency space (ELCFS).
	The distribution information is exchanged across clients from frequency space with an continous interpolation mechanism, enabling each local client to access the multi-source distributions. An episodic training paradigm is then established to expose the local optimization to domain shift, with explicit regularization to promote  domain-independent feature cohesion and separation at the ambiguous boundary region for improving generalizability.}
% 	\caption{Overview of our proposed method for federated domain generalization on medical image segmentation.
% 	The distribution information is exchanged across clients from frequency space with an interpolation mechanism, enabling local client access the multi-source distributions in a continuous space. An episodic training paradigm is then established to expose the local optimization to transferred domain shift, with explicit regularization to promote  domain-independent feature cohesion and separation in the ambiguous boundary region.}
	\label{fig:method}
	\vspace{-1.5mm}
\end{figure*}
\section{Method}
%We first describe the problem setting of federated domain generalization, and analyze its challenges in medical image segmentation scenario.
We start with the formulation for federated domain generalization and  its challenges in medical image segmentation scenario. We then describe the proposed method \textit{Episodic Learning in Continuous Frequency Space} (ELCFS) to explicitly address these challenges. An overview of the method is shown in Fig.~\ref{fig:method}.

% We then describe the proposed method, with an overview shown in Fig.~\ref{fig:method}.
% Specifically, we first present a privacy-preserving way leveraging frequency space to enable local client to access multi-source distributions, and then establish an episodic learning scheme with emphasizing regularization on ambiguous boundary region to learn generalizable FL models for medical image segmentation problem. 

\subsection{Federated Domain Generalization}
\label{formulation}
\textbf{Preliminaries:} 
In FedDG, we denote ($\mathcal{X}, \mathcal{Y}$) as the joint image and label space of a task, $\mathcal{S}=\{\mathcal{S}^1,\mathcal{S}^2,...,\mathcal{S}^K\}$ as the set of $K$ distributed source domains involved in federated learning. Each domain contains data and label pairs of $\mathcal{S}^k=\{(x^{k}_i,y^{k}_i)\}_{i=1}^{N^k}$, which are sampled from a domain-specific distribution $(\mathcal{X}^k, \mathcal{Y})$.
The goal of FedDG is to learn a model $f_\theta: \mathcal{X} \rightarrow \mathcal{Y}$ using the $K$ distributed source domains, such that it can directly generalize to a completely unseen testing domain $\mathcal{T}$ with a high performance.

Standard federated learning paradigm involves the communication between a central server and the $K$ local clients. At each federated round $t$, every client $k$ will receive the same global model weights $\theta$ from the central server and update the model with their local data $\mathcal{S}^k$ for $E$ epochs.
The central server then collects the local parameters $\theta^k$ from all clients and aggregates them to update the global model. This process repeats until the global model converges. In this work, we consider the most popular federated averaging algorithm (FedAvg)~\cite{mcmahan2017communication}, which aggregates the local parameters with weights in proportional to the size of each local dataset to update the global model, i.e., $\theta = \sum_{k=1}^K \frac{N^k}{N} \theta^k$, where $N = \sum_{k=1}^K N^k$. It is worth noting that our method can also be flexibly incorporated to other FL backbones.
%
 %Detailed procedure are listed in Algorithm xx.
% ~\textbf{Challenges:} The mainstream 

\textbf{Challenges:} 
With the goal of unseen domain generalization, a model is expected to thoroughly investigate the multi-source data distributions to pursue domain-invariance of its learned latent space. However, the federated setting in the specific medical image segmentation scenario poses several challenges for that.
\textit{First}, the multi-source data in FL are stored distributedly and the learning at each client can only access its individual local distribution, which constrains to make full use of the multi-source distributions to learn generalizable parameters. 
\textit{Second}, though FL has collaborated multi-source data, the medical images acquired from different clinical sites can present large heterogeneity. This leads to distinct distributions among the collaborative datasets, which is insufficient to ensure domain invariance in a more continuous distribution space to attain good generalizability in complex clinical environments. %Model trained in this way still suffer from inferior generalizability in complex clinical environments.
\textit{Third}, the structure of medical anatomises usually
present high ambiguity around its boundary region, raising challenge for previous DG techniques that typically lacks assurance for the domain-invariance of features in such ambiguous region.
\subsection{Continuous Frequency Space Interpolation }
\label{fsm}
To address the restriction of decentralized datasets, the foundation of our solution is to exchange the distribution information across clients, such that each local client can get access to multi-source data distributions for learning generalizable parameters.
Considering that sharing raw images is forbidden, we propose to exploit the information inherent in the frequency space, which enables to separate the distribution (i.e. style) information from the original images to be shared between clients without privacy leakage.

Specifically, given a sample $x_i^k \in \mathbb{R}^{H \times W \times C}$ ($C=3$ for RGB image and $C=1$ for grey-scale image) from the $k$-th client, we can obtain its frequency space signal through fast Fourier transform~\cite{nussbaumer1981fast} as:
\begin{equation}
\small
\mathcal{F}(x_i^k)(u,v,c) = \sum_{h=0}^{H-1} \sum_{w=0}^{W-1} x_{i}^k(h,w,c) e^{-j2\pi(\frac{h}{H}u+\frac{w}{W}v)}.
\end{equation}
This frequency space signal $\mathcal{F}(x_i^k)$ can be further decomposed to an amplitude spectrum $\mathcal{A}_i^{k} \in \mathbb{R}^{H\times W\times C}$ and a phase spectrum $\mathcal{P}_i^{k} \in \mathbb{R}^{H\times W\times C}$, which respectively reflect the low-level distributions (e.g. style) and high-level semantics (e.g. object) of the image. 
To exchange the distribution information across clients, we first construct a distribution bank $\mathcal{A} = [\mathcal{A}^{1},..., \mathcal{A}^{K}]$, where each $\mathcal{A}^{k} = \{\mathcal{A}_i^{k}\}_{i=1}^{N^k}$ contains all amplitude spectrum of images from the $k$-th client, representing the distribution of $\mathcal{X}^k$.
This bank is then made accessible to all clients as shared distribution knowledge.

%Next, we propose to exchange the amplitude spectrum across clients to transmit the distribution information

Next, we design a continuous interpolation mechanism within the frequency space, aiming to transmit multi-source distribution information to a local client leveraging the distribution bank.
As shown in the left part of Fig.~\ref{fig:method}, given a local image $x_i^k$ at client $k$, we can replace some low-frequency component of its amplitude spectrum with the ones in distribution bank $\mathcal{A}$, while its phase spectrum is unaffected to preserve the semantic content. As an outcome, we can generate images with transformed appearances exhibiting distribution characteristics of other clients. More importantly, we continuously interpolate between amplitude spectrum of local data and the transferred amplitude spectrum of other domains.
In this way, we can enrich the established multi-domain distributions for each local client, benefiting from a dedicated dense space with smooth distribution changes.
%so that it can be transformed to other distributions (i.e. style) by replacing the low-frequency component of its amplitude spectrum with the one containing desired distribution information, while with the phase spectrum fixed to preserve the semantic content. 
%\red{Based on this, to transform local data $x_i^k$ to the interacted distribution space between $\mathcal{X}^{k}$ and $\mathcal{X}^{n}$, we need to synthesize the particular amplitude spectrum that contains the interacted distribution information.}
Formally, this is achieved by randomly sampling an amplitude spectrum item $\mathcal{A}_j^{n} (n\neq k)$ from the distribution bank, then synthesize a new amplitude spectrum by interpolating between $\mathcal{A}_i^{k}$ and $\mathcal{A}_j^{n}$. Let $\mathcal{M}=\mathbbm{1}_{(h, w)\in[-\alpha H: \alpha H, -\alpha W: \alpha W]}$ be a binary mask which controls the scale of low-frequency component within amplitude spectrum to be exchanged, whose value is 1 at the central region and 0 elsewhere. Denote $\lambda$ as the interpolation ratio adjusting the amount of distribution information contributed by $\mathcal{A}_i^{k}$ and $\mathcal{A}_j^{n}$, the generated new amplitude spectrum interacting distributions for local client $k$ and external client $n$ is represented as:
\begin{equation}
\small
\mathcal{A}_{i,\lambda}^{k \to n}  = (1-\lambda)\mathcal{A}_i^{k}\ast (1-\mathcal{M}) + \lambda\mathcal{A}_j^{n}\ast \mathcal{M}.
\label{mixup}
\end{equation}
%where $\lambda$ is the interpolation ratio determining how much distribution information are involved from $\mathcal{A}_i^{k}$ and $\mathcal{A}_j^{n}$ respectively. 
%
After obtaining the interpolated amplitude spectrum $\mathcal{A}_{i,\lambda}^{k \to n}$, we then combine it with the original phase spectrum to generate the transformed image via inverse Fourier transform $\mathcal{F}^{-1}$ as follows:
\begin{equation}
\small
x_{i,\lambda}^{k \to n} = \mathcal{F}^{-1} (\mathcal{A}_{i,\lambda}^{k \to n} , \mathcal{P}_i^{k}),
\label{inverseFFT}
\end{equation}
where the generated image $x_{i,\lambda}^{k \to n}$ preserves the original semantics of $x_i^k$ while carrying a new distribution interacted between $\mathcal{X}^k$ and $\mathcal{X}^n$. In our implementation, the interpolation ratio $\lambda$ will be dynamically sampled from [0.0, 1.0] to generate images via a continuous distribution space. 
As intuitive examples shown in Fig.~\ref{fig:method}, our interpolation operation allows the generated samples to bridge the intermediate space between distinct distributions across domains.
Note that the method described above does not require heavy computations, thus can be performed online as the local learning goes on. Practically, for each input $x_i^k$, we will sample an amplitude spectrum $A_j^n$ from the distribution bank for each external client $n \neq k$, and transform its image appearance as Eqs.~(2-3). Through this, we obtain $K\!-\!1$ transformed images $\{x_{i,\lambda}^{k\rightarrow n}\}_{ n\neq k}$ of different distributions, which share the same semantic label as $x_i^k$.
%since the phase spectrum with semantic content is unchanged. 
For ease of denotation, we represent these transformed images as $t_i^k$ hereafter, i.e. $t_i^k$=$\{x_{i,\lambda}^{k\rightarrow n}\}_{ n\neq k}$.
Furthermore, this approach does not violate the privacy concern since the phase spectrum containing core semantics are retained at each client throughout the whole process, and the raw images cannot be reconstructed with the amplitude spectrum alone~\cite{schomberg1995gridding}.

\subsection{Boundary-oriented Episodic Learning}
\label{meta}

The above constructed continuous multi-source distributions at each local client provide the materials to learn generalizable local parameters. In the following, we carefully design a boundary-oriented episodic learning scheme for local training, by particularly meeting challenges of model generalization in medical image segmentation scenario.
%The above constructed multi-source distributions from frequency space provide the materials for local client to learn generalizable parameters. Whereas to achieve this, the local learning scheme certainly requires to be carefully designed to fully exploit the distribution information, since the generalizable parameters is not naturally obtained. In the following, we carefully design a boundary-oriented episodic learning scheme for local training, by particularly meeting challenges in medical image segmentation scenario.

\textbf{Episodic learning at local client:}
We establish the local training as an episodic meta-learning scheme, which learns generalizable model parameters by simulating train/test domain shift explicitly. Note that in our case, the domain shift at a local client comes from the data generated from frequency space with different distributions.  
Specifically, in each iteration, we consider the raw input $x_i^k$ as meta-train and its counterparts $t_i^k$ generated from frequency space as meta-test presenting distribution shift (cf. Fig.~\ref{fig:method}). The meta-learning scheme can then be decoupled to two steps. First, the model parameters $\theta^k$ are updated on meta-train
with segmentation Dice loss $\mathcal{L}_{seg}$:
\begin{equation}
\small
\hat{\theta}^{k}= \theta^k - \beta \nabla_{\theta^k}\mathcal{L}_{seg}(x_i^k; \theta^k),
\label{innerloop}
\end{equation}
where $\beta$ denotes the learning rate for the inner-loop update. Second, a meta-update is performed to virtually evaluate the updated parameters $\hat{\theta}^{k}$ on the held-out meta-test data $t_i^k$ with a meta-objective $\mathcal{L}_{meta}$. Crucially, this objective is computed with the updated parameters $\hat{\theta}^{k}$, but optimized w.r.t the original parameters $\theta^k$. Such optimization paradigm aims to train the model such that its learning on source domains can further fulfill certain properties that we desire in unseen domains, 
which are quantified by $\mathcal{L}_{meta}$.

\textbf{Boundary-oriented meta optimization:} We define the $\mathcal{L}_{meta}$ with considering specific challenges in medical image segmentation. Particularly, it is observed that the performance drop of segmentation results at unseen domains outside federation often comes from the ambiguous boundary area of anatomies. %Guiding the model to improve domain-invariant discrimination capability for pixels around the boundary is crucial for the generalization problem in medical image segmentation. 
To this end, we design a new boundary-oriented objective to enhance the domain-invariant boundary delineation, by carefully learning from the local data $x_i^k$ and the corresponding $t_i^k$ generated from frequency space with multi-source distributions. The idea is to regularize the boundary-related and background-related  features of these data to respectively cluster to a compact space regardless of their distributions while reducing the clusters overlap.
This is crucial, since if the model cannot project their features around boundary area with distribution-independent class-specific cohesion and separation, the predictions will suffer from ambiguous decision boundaries and still be sensitive to the distribution shift when deployed to unseen domains outside federation.
%We define the $\mathcal{L}_{meta}$ with considering specific challenges in medical image segmentation. Particularly, it is observed that the performance drop of segmentation results at unseen domains outside federation often comes from the ambiguous boundary area of anatomies. Guiding the model to improve domain-invariant discrimination capability for pixels around the boundary by effectively exploiting the diverse distribution information from frequency space is crucial for the generalization problem in medical image segmentation. To this end, we design a new boundary-oriented objective $\mathcal{L}_{boundary}$ in meta-optimization, to regularize the boundary-related and background-related  features to respectively cluster to a compact feature space regardless of the distributions transferred from frequency space while reducing their overlap. This is crucial, since if the model cannot project features around boundary area with domain-independent class-specific cohesion and separation, the predictions will suffer from ambiguous decision boundaries and still be sensitive to the distribution shift when deployed to unseen domains outside participanting clients.

Specifically, we first extract the boundary-related and background-related features for the input samples. Given image $x_i^k$ with segmentation label $y_i^k$,  we can extract its binary boundary mask $y_{i\_bd}^k$ and background mask $y_{i\_bg}^k$ with morphological operations on $y_i^k$.
Here, the mask $y_{i\_bg}^k$ only contains background pixels around the anatomy boundary instead of from the whole image, as we expect to enhance the discriminability for features around the boundary region. Let $Z_i^k$ denote the activation map extracted from layer $l$, which is interpolated with  bilinear interpolation to keep consistent dimensions as $y_i^k$. Then the boundary-related and background-related features of $x_i^k$ can be extracted from $Z_i^k$ with masked average pooling over $y_{i\_bd}^k$ and $y_{i\_bg}^k$ as:
\begin{equation}
\small
h_{i\_bd}^k=\frac{\sum_{h,w}  Z^k_i \ast y_{i\_bd}^k}{\sum_{h,w} y_{i\_bd}^k};
h_{i\_bg}^k=\frac{\sum_{h,w}  Z^k_i  \ast y_{i\_bg}^k}{\sum_{h,w} y_{i\_bg}^k},
\label{eq:embeddings}
\end{equation}
where $\ast$ denote element-wise product. The produced $h_{i\_bd}^k$ and $h_{i\_bg}^k$ are single-dimensional vectors, representing the averaged region-level features of the boundary and background pixels.
By further performing the same operation for $K$-$1$ transformed images $t_i^k$  with different distributions transferred from the frequency space, we accordingly obtain together $K$ boundary-related and $K$ background-related features. 

Next, we enhance the domain-invariance and discriminability of these features, by regularizing their intra-class cohesion and inter-class separation regardless of distributions.   
Here, we employ the well-established  InfoNCE~\cite{chen2020simple} objective to impose such regularization.
Denote ($h_m, h_p$) as a pair of features, which is a positive pair if $h_m$ and $h_p$ are of the same class (both boundary-related or background-related) and otherwise negative pair. 
In our case, the InfoNCE loss is defined over each positive pair ($h_m, h_p$) within the $2 \times K$ region-level features as:
% \begin{equation}
% \small
% \ell(h_l,h_m)=-log\frac{exp(h_l \odot h_m / \tau)}{\sum_{q=1}^{2K} \mathbb{F}(h_l, q)\cdot exp(h_l \odot h_m / \tau)},
% \end{equation}
\begin{equation}
\small
\ell(h_m,h_p)=-log\frac{exp(h_m \odot h_p / \tau)}{\sum_{q=1,q\neq m}^{2K} \mathbb{F}(h_m, h_q)\cdot exp(h_m \odot h_q / \tau)},
\end{equation}
where $\odot$ denote the cosine similarity: $a \odot b = \frac{\langle a,b \rangle}{||a||_2||b||_2}$; the value of $\mathbb{F}(h_m, h_q)$ is 0 and 1 for positive and negative pair respectively; $\tau$ denotes the temperature parameter. The final loss $\mathcal{L}_{boundary}$ is the average of $\ell$ over all positive pairs:
\begin{equation}
\small
\mathcal{L}_{boundary}=\sum_{m=1}^{2K}\sum_{p=m+1}^{2K}\frac{(1 - \mathbb{F}(h_m, h_p))\cdot \ell (h_m, h_p)}{B(K,2)\times2},
\end{equation}
where $B(K,2)$ is the number of combinations.

\textbf{Overall local learning objective:}
The overall meta objective is composed of the segmentation dice loss $\mathcal{L}_{seg}$ and the boundary-oriented objective $\mathcal{L}_{boundary}$ as:
\begin{equation}
\mathcal{L}_{meta} = \mathcal{L}_{seg}(t_i^k;\hat{\theta}^{k}) + \gamma\mathcal{L}_{boundary}(x_i^k, t_i^k;\hat{\theta}^{k}),
\end{equation}
where $\hat{\theta}^k$ is the updated parameter from Eq.~\ref{innerloop}, $\gamma$ is a balancing hyper-parameter. Finally, both the inner-loop objective and  meta objective will be optimized together with respect to the original parameter $\theta^k$ as:
\begin{equation}
\mathop{\arg\min}_{\theta^k} \mathcal{L}_{seg}(x_i^k; \theta^k) + \mathcal{L}_{meta}(x_i^k, t_i^k;\hat{\theta}^{k}).
\label{eq:meta}
\end{equation}
In a federated round, once the local learning is finished, the local parameters $\theta^k$ from all clients will be aggregated at the central server to update the global model.

\section{Experiments}
We extensively evaluate our method on two medical image segmentation tasks, i.e., the optic disc and cup segmentation on retinal fundus images~\cite{orlando2020refuge}, and the prostate segmentation on T2-weighted MRI~\cite{litjens2014evaluation}.
We first conduct comparison with DG methods that can be incorporated in the federated paradigm, and then provide in-depth ablation studies to analyze our method.
% Further, we conduct in-depth ablation studies to analyze the contribution of each key component, the importance of the continuous interpolation, the discriminability at ambiguous boundary region, and the effect of participating client number.

% For each task, we collect data from multiple different clinical sites out of public datasets to simulate the federated learning scenario. The data acquired from different sources present heterogeneous distributions due to the different scanning protocols used for image acquisition. 
\subsection{Datasets and Evaluation Metrics}
\begin{figure}[t]
	\centering
	\includegraphics[width=0.94\textwidth]{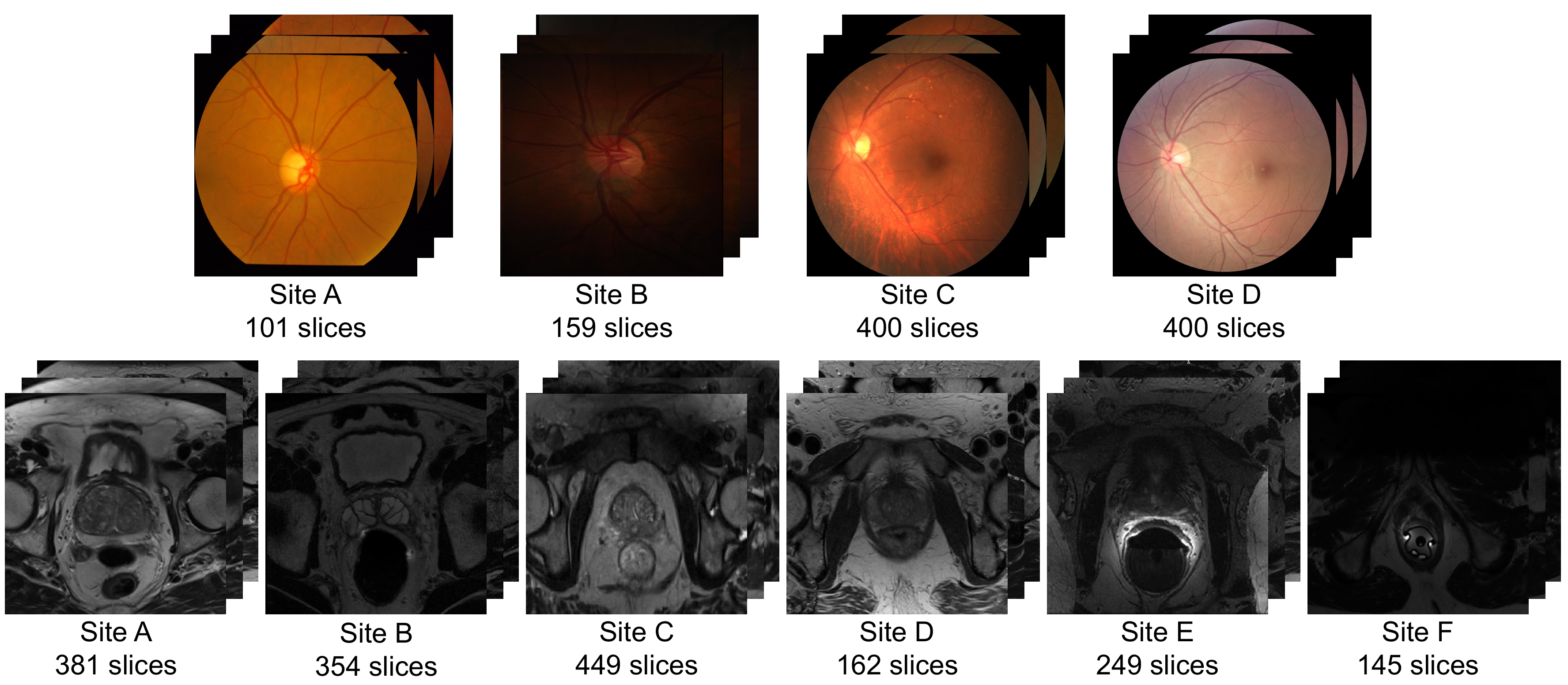}
% 	\caption{Fundus images (first row) and prostate MRIs (second rwo) acquired from different clinical sites, presenting heterogeneous appearances due to various imaging scanners and protocols.} 
    \caption{Example cases and slice number of each data source in fundus image segmentation and prostate MRI segmentation tasks.} 
 	\label{fig:statistics}
 	\vspace{-1.5mm}
\end{figure}

We employ \textbf{retinal fundus images from 4 different clinical centers} of public datasets~\cite{sivaswamy2015comprehensive,fumero2011rim,orlando2020refuge} for optic disc and cup segmentation. For pre-processing, we center-crop a $800 \! \times \! 800$ disc region for these data uniformly, then resize the cropped region to $384 \! \times \! 384$ as network input. We further collect \textbf{prostate T2-weighted MRI images from 6 different data sources} partitioned from the public datasets~\cite{bloch2015nci,lemaitre2015computer,litjens2014evaluation,liu2020ms} for prostate MRI segmentation task. All the data are pre-processed to have similar field of view for the prostate region and resized to $384 \! \times \! 384$ in axial plane. We then normalize the data individually to zero mean and unit variance in intensity values.    %The typical examples and sample numbers of each data source are presented in Fig.~\ref{fig:statistics}. 
Note that for both tasks, the data acquired from different clinical centers present heterogeneous distributions due to the varying imaging conditions. The example cases and sample numbers of each data source are presented in Fig.~\ref{fig:statistics}. Data augmentation of random rotation, scaling, and flipping are employed in the two tasks. For evaluation, we adopt two commonly-used metrics of Dice coefficient (Dice) and Hausdorff distance (HD), to quantitatively evaluate the segmentation results on the whole object region and the surface shape respectively. %More details about the datasets are shown in Appendix. %A higher Dice value and lower Hausdorff distance indicate a better segmentation results. 

\subsection{Implementation Details}
% We design specific task models based on Mix-residual-UNet~\cite{yu2017volumetric} for the two segmentation tasks, please refer to the supplementary material for more details. Due to the large variance on slice thickness of prostate MRI data, we employ the 2D architecture for the two tasks. %
In the federated learning process, all clients use the same hyper-parameter settings, and the local model is trained using Adam optimizer with batch size of 5 and Adam momentum of 0.9 and 0.99. The meta-step size and learning rate are both set as $1e^{-3}$. 
The interpolation ratio $\lambda$ in frequency space is randomly sampled within [0.0, 1.0], and we will investigate this parameter in the ablation study. The hyper-parameter $\alpha$ is empirically set as 0.01 to avoid artifacts on the transformed images. The activation map from the last two deconvolutional layers are interpolated and concatenated to extract the semantic features around boundary region, and the temperature parameter $\tau$ is empirically set as 0.05. The weight $\gamma$ is set as 0.1 and 0.5 in the two tasks to balance the magnitude of the training objectives. 
We totally train 100 federated rounds as the global model has converged stably, and the local epoch $E$ in each federated round is set as 1.
The framework is implemented with Pytorch library, and is trained on two NVIDIA TitanXp GPUs. %The computation for each client is conducted on one GPU. 

\begin{table*}[!tbp]
    \renewcommand\arraystretch{1.2}
    \centering
        \caption{\small{Comparison of federated domain generalization results on Optic Disc/Cup segmentation from fundus images.}}
        \label{tab:comparisonsfundus}
        \vspace{-2mm}
        \resizebox{1.0\textwidth}{!}{%
        \setlength\tabcolsep{3.0pt}
        \scalebox{0.69}{
        \begin{tabular}{c|c  c  c  c c| c cccc |c||c  c  c  c c| c cccc |c}
    %       \hline
            \hline
            \hline
             Task&\multicolumn{5}{c|}{Optic Disc Segmentation} &\multicolumn{5}{c|}{Optic Cup Segmentation} &\multirow{2}{*}{Overall} &\multicolumn{5}{c|}{Optic Disc Segmentation} &\multicolumn{5}{c|}{Optic Cup Segmentation} &\multirow{2}{*}{Overall}\\
            \cline{0-10}
            \cline{13-22}
            
            Unseen Site& A & B & C & D &Avg. & A & B & C & D &Avg. & & A & B & C & D &Avg. & A & B & C & D &Avg. &\\
            \hline
            \hline
            &\multicolumn{11}{c||}{\textbf{Dice Coefficient (Dice)~$\uparrow$}} &\multicolumn{11}{c}{\textbf{Hausdorff Distance (HD)~$\downarrow$}}\\ 
            \hline
            % Shared data &94.89$\pm$2.47 &88.82$\pm$8.22&92.96$\pm$3.09 &92.40$\pm$3.49 &92.27 &82.60$\pm$10.35&71.60$\pm$19.20 &83.03$\pm$8.76 &83.84$\pm$14.74 &80.27 \\
            \hline
            
            JiGen~\cite{carlucci2019domain}  &93.92 &85.91 &92.63 &94.03 &91.62 &82.26 &70.68 &83.32 &\textbf{85.70} &80.47 &86.06 &13.12 &20.18 &11.29 &8.15 &13.19 &20.88 &23.21 &11.55 &9.23 &16.22 &14.71\\
            BigAug~\cite{zhang2020generalizing}  &93.49 &86.18 &92.09 &93.67 &91.36 &81.62 &69.46 &82.64  &84.51 &79.56 &85.46 &16.91 &19.01 &11.53 &8.76 &14.05 &21.21 &23.10 &12.02 &10.47 &16.70&15.39\\
            Epi-FCR~\cite{li2019episodic} &94.34 &86.22 &92.88 &93.73 &91.79 &83.06&70.25&83.68 &83.14 &80.03 &85.91 &13.02 &18.97 &\textbf{10.67}&8.47 &12.78 &19.12&21.94& 11.50 &10.86 &15.86&14.32\\
            RSC~\cite{huang2020self}&94.50 &86.21 &92.23 &94.15 &91.77 &81.77 &69.37 &83.40 &84.82 &79.84 &85.80 &19.44 &19.26&13.47 &8.14 &15.08 &23.85 &24.01 &11.38 &9.79 &17.25 &16.16\\ 
            % Fed-Epi-FCR~\cite{huang2020self}\\  
            % \textbf{Fed + frequency (I)} &94.03 &86.64 &92.68 92.79 &93.56 &91.76\\
            % \textbf{Fed + mixup (II)} &94.16 &87.29 &92.70 &94.47 &92.15\\
            % \textbf{Fed + meta (III)}&\textbf{94.37$\pm$3.24} &\textbf{87.64$\pm$6.47} &\textbf{93.37$\pm$3.38} &\textbf{94.50$\pm$2.80} &\textbf{92.47} &\textbf{82.77$\pm$11.74} &70.54$\pm$19.40 &\textbf{83.94$\pm$8.63} &\textbf{85.51$\pm$8.45} &\textbf{80.64} \\
            \hline
            FedAvg~\cite{mcmahan2017communication} &92.88 &85.73 &92.07 &93.21 &90.97 &80.84 &69.71 &82.28 &83.35 &79.05 &85.01 &17.01 &20.68 &11.70 &9.33 &14.68 &20.77 &26.01 &11.85 &10.03 &17.17 &15.93\\
            \textbf{ELCFS (Ours)} &\textbf{95.37} &\textbf{87.52} &\textbf{93.37} &\textbf{94.50} &\textbf{92.69} &\textbf{84.13}  &\textbf{71.88} &\textbf{83.94} &85.51 &\textbf{81.37} &\textbf{87.03} &\textbf{11.36} &\textbf{17.10} &10.83 &\textbf{7.24}&\textbf{11.63} &\textbf{18.65} &\textbf{19.36} &\textbf{11.17} &\textbf{8.91} &\textbf{14.52} &\textbf{13.07}\\
            \hline
            \hline
            
            % &\multicolumn{10}{c}{\textbf{Hausdorff Distance (mean$\pm$std)~$\downarrow$}} \\ 
        \end{tabular}

    }}
    \vspace{-1mm}
\end{table*}

\begin{table*}[!tbp]
    \renewcommand\arraystretch{1.2}
    \centering
        \caption{\small{Comparison of federated domain generalization results on prostate MRI segmentation.}}
                \vspace{-2mm}
        \resizebox{0.8\textwidth}{!}{%
        \setlength\tabcolsep{5.0pt}
        \scalebox{0.73}{
        \begin{tabular}{c|c  c  c  c c c c||c  c  c  c c c c}
    %       \hline
            \hline
            \hline
            Unseen Site & A & B &  C & D & E & F &Average & A & B &  C & D & E & F &Average\\
            \hline
            \hline
            &\multicolumn{7}{c||}{\textbf{Dice Coefficient (Dice)~$\uparrow$}} &\multicolumn{7}{c}{\textbf{Hausdorff Distance (HD)~$\downarrow$}}\\ 
            \hline
            \hline
            JiGen~\cite{carlucci2019domain} &89.95 &85.81 &84.06 &87.34 &81.32 &89.11 &86.26 &10.51 &11.53 &11.70 &11.49 &14.80 &9.02 &11.51\\
            BigAug~\cite{zhang2020generalizing} &89.63 &84.62 &83.86 &87.66 &81.20 &88.96 &85.99 &10.68 &11.78 &12.07 &\textbf{10.66} &13.98 &9.73 &11.48\\ 
            Epi-FCR~\cite{li2019episodic}&89.72 &85.39 &84.97 &86.55 &80.63 &89.76 &86.17&10.60 &12.31 &12.29 &12.00 &15.68 &8.81 &11.95\\
            RSC~\cite{huang2020self} &88.86 &85.56 &84.36 &86.21 &79.97 &89.80 &85.80 &10.57 &11.84 &14.76 &13.07 &14.79 &8.83 &12.31\\  
            \hline
            FedAvg~\cite{mcmahan2017communication} &89.02 &84.48 &84.11 &86.30 &80.38 &89.15 &85.57 &11.64 &12.01 &14.86 &11.80 &14.90 &9.30 &12.42\\
            \textbf{ELCFS (Ours)} &\textbf{90.19} &\textbf{87.17} &\textbf{85.26} &\textbf{88.23} &\textbf{83.02} &\textbf{90.47} &\textbf{87.39} &\textbf{10.30} &\textbf{11.49} &\textbf{11.50} &11.57 &\textbf{11.08} &\textbf{8.31} &\textbf{10.88}\\
            \hline
            \hline

        \end{tabular}

    }}
\label{tab:comparisons_prostate}
\vspace{-3mm}
\end{table*}

\subsection{Comparison with DG methods}

\textbf{Experimental setting:} In our experiments, we follow the practice in domain generalization literature to adopt the leave-one-domain-out strategy, i.e., training on $K$-$1$ distributed source domains and testing on the one left-out unseen target domain. This results in four generalization settings for the  fundus image segmentation task and six settings for the prostate MRI segmentation task.

% Since many domain generalization methods require access to multi-domain data in the learning process, applying them in federated setting is less feasible. 
% We therefore compare with other domain generalization approaches that do not rely on data centralization and can be incorporated into the local learning process in federated scenario. 

% \vspace{-10.5mm}
We compare with recent state-of-the-art DG methods that are free from data centralization and can be incorporated into the local learning process in federated paradigm, including:
\textbf{JiGen~\cite{carlucci2019domain}} an effective self-supervised learning approach to learn general representations by solving jigsaw puzzles;
\textbf{BigAug~\cite{zhang2020generalizing}} a method that performs extensive data transformations to regularize general representation learning;
\textbf{Epi-FCR~\cite{li2019episodic}} a scheme to periodically exchange partial model (classifier or feature extractor) across domains to expose model learning to domain shift;
\textbf{RSC~\cite{huang2020self}} a method that randomly discards the dominating features to promote robust model optimization.
% For the implementation of BigAug, we follow the authors to add nine kinds of data transformation techniques in the learning process. For the other three methods, we modify the code provided by authors to establish them in the federated setting.
For the implementation, we follow their public code or paper and establish them in the federated setting.
We also compare with the baseline setting, i.e., learning a global model with the basic \textbf{FedAvg~\cite{mcmahan2017communication}} algorithm without any generalization technique.

\begin{figure}[t]
	\centering
	\includegraphics[width=0.96\textwidth]{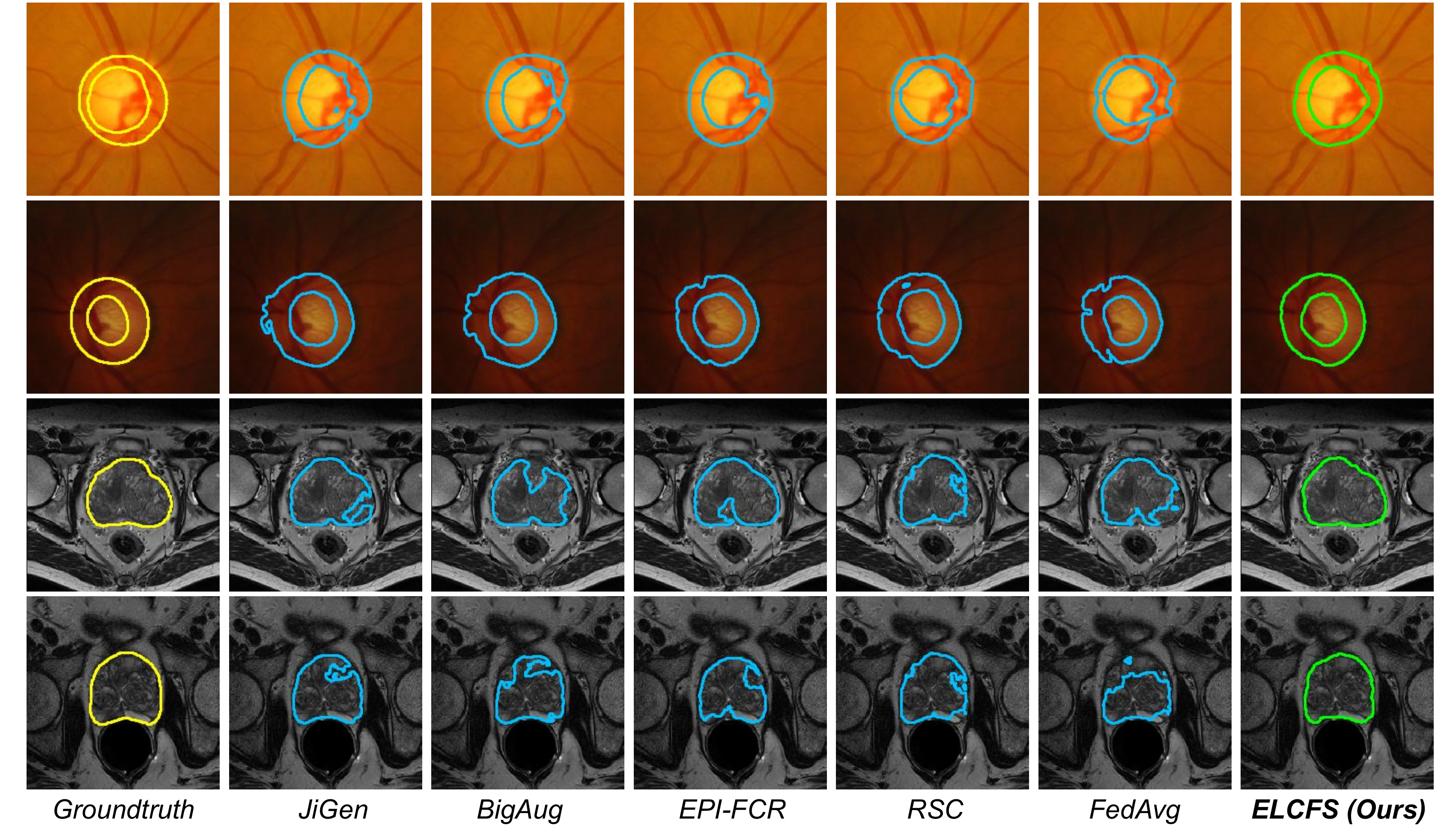}
	\caption{Qualitative comparison on the generalization results of different methods in fundus image segmentation (top two rows) and prostate MRI segmentation (bottom two tows).} 
 	\label{fig:visualcompare}
 	 		 \vspace{-1.5mm}
\end{figure}

\textbf{Comparison results:}
Table~\ref{tab:comparisonsfundus} presents the quantitative results for retinal fundus segmentation.
% It is observed that BigAug obtains the least improvement over the FedAvg baseline, which may because data augmentations have already been conducted in the data preprocessing, making the benefits of including extra transformations less obvious.
% The other three methods Fed-RSC, Fed-Epi-FCR, and Fed-JiGen which aim to learn more generalizable feature representations in the local distributions, achieve similar generalization performance
We see that different DG methods can improve the overall generalization performance more or less over FedAvg. This attributes to their regularization effect on the local learning to extract general representations. 
% Compared with these methods, our approach achieves general improvements on most unseen sites for Dice and HD on optic disc and cup segmentation.
% Our approach achieves the highest overall performance and the best results on most unseen sites in terms of Dice and HD for optic disc and cup segmentation.
Compared with these methods, our ELCFS achieves higher overall performance and obtains improvements on most unseen sites in terms of Dice and HD for both optic disc and cup segmentation.
%on all unseen sites for optic disc segmentation and site A/B/D for optic cup segmentation. 
This benefits from our frequency space interpolation mechanism which presents multi-domain distributions to local client.
Specifically, for other DG methods, their local learning still can only access the individual distribution and fail to regularize the features towards domain-invariance in a diverse distribution space. In contrast, our method enables the local learning to take full advantages of the multi-source distributions and explicitly enhances the domain-invariance of features around the ambiguous boundary region. 
% In addition, the proposed method achieves consistent improvements over FedAvg across all unseen domain settings, with increase of 2.02\% (87.03\%-85.01\%) in Dice and 2.86 (15.93-13.07) in HD for the overall performance.  
In addition, our ELCFS achieves consistent improvements over FedAvg across all unseen domain settings, with the overall performance increase of 2.02\% in Dice and 2.86 in HD.  
%Whereas for other methods, the local learning still can only access the individual distribution and fail to regularize the features towards domain-invariance in a wide distribution space.    
% Whereas in these methods, the local learning still can only access the individual distribution, failing to comprehensively exploit of multi-source distributions to attain a higher generalization performance. In contrast, our method, by enabling local client to make fully use of the joint distribution information from different sources, achieves the highest performance on all the sites for optic disc segmentation, and the best results on site A/B/C with comparable results on site D for optic cup segmentation. The proposed method also considerably outperforms the FedAvg baseline model by 2.02\% in the overall Dice score and 2.86 in the Hausdorff distance. 
% For prostate MRI segmentation, the best performance among the four comparison methods is 86.26\% in Dice (JiGen) and 11.48 in HD (BigAug), which is higher than the baseline FedAvg method. In contrast, our method can obtain a better average performance on the two metrics, with consistent Dice improvements over these methods across all six sites. 
For prostate MRI segmentation, the comparison DG methods generally perform better than FedAvg, but the improvements are relatively marginal. 
Our ELCFS obtains the highest Dice across all the six unseen sites and HD on most sites. 
Overall, our method improves over FedAvg for Dice from 85.57\% to 87.39\% and HD from 12.42 to 10.88, outperforming other DG methods. %The standard deviation of the comparison results is included in Appendix due to page limit. 
Fig.~\ref{fig:visualcompare} shows the segmentation results with two cases from unseen domains for each task. It is observed that our method accurately segments the structure and delineates the boundary in images of unknown distributions, whereas other methods sometimes fail to do so.

\subsection{Ablation Analysis of Our Method}
We conduct ablation studies to investigate four key questions regarding our ELCFS: \textbf{1)} the contribution of each component to our model performance, \textbf{2)} the benefit of the interpolation operation and the choice of $\lambda$, \textbf{3)} how the semantic feature space around the boundary region is influenced by our method, and \textbf{4)} how the numbers of participanting clients affect the performance of our method.

\begin{figure}[t]
	\centering
	\includegraphics[width=1\textwidth]{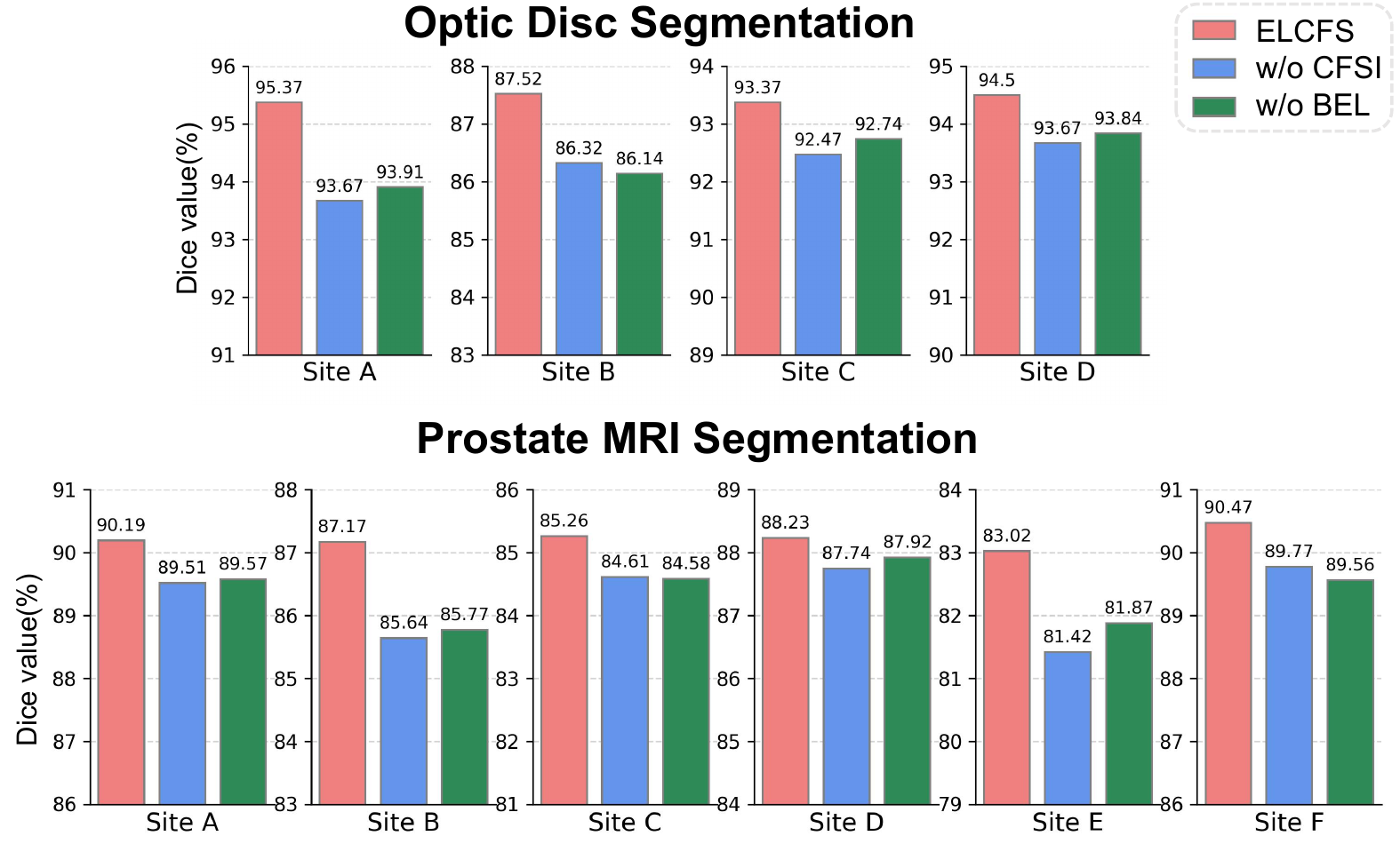}

	\caption{Ablation results to analyze the effect of the two components (i.e. CFSI and BEL) in our method.}
% 	\caption{Ablation results on the two tasks. The red, blue, and green bars present the performance of our full model, removing CFSI, and removing BEL respectively. The dotted line denote baseline FedAvg model.}
% 	\caption{Ablation results on the two tasks. The pink bars are the generalization dice score of our method, and the blue, green and brown bars denote the performance when removing the frequency interpolation, meta-learning, and ambiguity enhancement from our method. The black dotted line denote baseline performance without using generalization technique, i.e., pure FedAvg algorithm.} 
 	\label{fig:component}
 		 \vspace{-1.5mm}
\end{figure}

\textbf{Contribution of each component:}
We first validate the effect of the two key components in our method, i.e. continuous frequency space interpolation (\textbf{CFSI}) and Boundary-oriented Episodic Learning (\textbf{BEL}), by removing them respectively from our method to observe the model performance. As shown in Fig.~\ref{fig:component}, removing either part will lead to decrease on the generalization performance in different unseen domain settings for the two tasks. This is reasonable and reflects how the two components play complementary roles to the performance of our method, i.e., the generated distributions from CFSI lays foundation for the learning of BEL, and the BEL inversely provides assurance to effectively exploit the generated distributions. 
% both parts are indispensable to ensure a good generalization performance of our method. 

% compared with the full model, removing either component consistently leads to a decrease in Dice for each site on both segmentation tasks, demonstrating the contribution of CFSI and BEL to improving the generalization performance of the full model. 
% We also observe that, both the models ``w/o CFSI" and ``w/o BEL" increase the segmentation results over the baseline FedAvg on each generalization site, showing that either component 

% We also observe that after removing one component, both the models ``w/o CFSI" and ``w/o BREL" just obtain limited improvements over the FedAvg baseline model. This shows that the two components are mutually reinforcing each other and  equally important for federated domain generalization. 
% We also observe that removing the distribution simulation results in the most significant performance decrease on almost all the generalization settings. 
% For example in site A of fundus image segmentation, Dice score greatly drops from 95.4\% to 93.5\% when comparing the ``w/o DS" model with the full model.
% This indicates that simulating diverse multi-source distributions in each client is critical for the local training to learn updates with domain invariance, which can then be aggregated towards a more generalizable global model. 

\begin{figure}[t]
	\centering
	\includegraphics[width=1.0\textwidth]{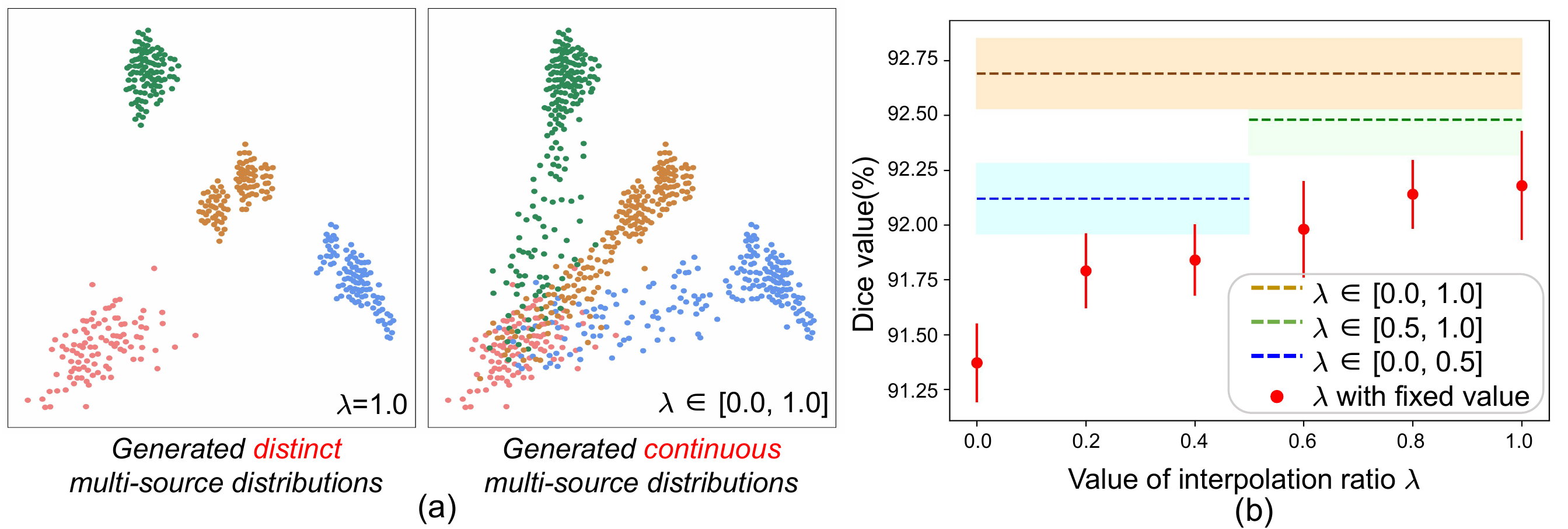}
	\caption{(a) Visualization of t-SNE~\cite{maaten2008visualizing} embedding for the original fundus images at a local client (pink points) and the corresponding transformed images with amplitude spectrum from different clients (green, yellow, and blue points); (b) Generalization performance on optic disc segmentation under different settings of interpolation ratio $\lambda$, with fixed value or continuous sampling from different ranges (with error bar from three independent runs).} 
% 	\caption{Analysis of continuous frequency space interpolation. (a) visualization of t-SNE~\cite{maaten2008visualizing} embedding for interpolated images between domains (using fundus dataset); (b) model generalization performance under different settings of interpolation ratio $\lambda$ with fixed value and continuous interpolation in different ranges (error bar from three independent runs).} 
% 	\caption{Analysis of continuous frequency space interpolation. (a)-(c) The t-SNE embedding visualization of fundus images at a client. The red dots represent embeddings of original data. The dots of other colors represent embeddings of simulated data with the amplitude spectrum of different clients. In (b), data are simulated by directly swapping the amplitude spectrum between clients. In (c), data are simulated with our proposed continuous interpolation between the amplitude spectrum; (d) Comparison on averaged generalization performance of models with or without using the continuous interpolation; (e) Effect of the value of interpolation ratio $\lambda$ on the averaged generalization performance of OD segmentation, with error bar within three independent runs.} 
	\label{fig:continuous}
	 		 \vspace{-1.5mm}
\end{figure}

\textbf{Importance of continuous interpolation in frequency space:}
To analyze the effect of continuous interpolation mechanism in ELCFS , we use t-SNE~\cite{maaten2008visualizing} to visualize the distribution of generated images in fundus image segmentation. As shown in Fig.~\ref{fig:continuous} (a), the pink points denote the local data of a client, and other points denote the transformed data that are generated with amplitude spectrum from different clients. It appears that fixing $\lambda$ (left) will lead to several distinct distributions, while the continuous interpolation mechanism (right) can smoothly bridge the distinct distributions to enrich the established multi-domain distributions. This promotes the local learning to attain domain-invariance in a dedicted dense distribution space. 

% to harmonize the multi-source distribution more smoothly and bridge the gap enables the generated data to bride the distant distribution smoothly to harmonize the span over a continuous space enriching the multi-source distributions.
% Fig.~\ref{fig:continuous} (b) shows that directly swapping the amplitude spectrum of data samples between clients can generate multiple distinct distributions, demonstrating that the amplitude spectrum of frequency space signal contain important data statistics information.
% In Fig.~\ref{fig:continuous} (c), with our proposed continuous interpolation, we obtain continuous distributions spanning over the multi-source statistics.
% In Fig.~\ref{fig:continuous} (a) right, with our proposed continuous interpolation, the data embeddings become more continuous and less separable spanning over the multi-source distributions.
% We also quantitatively compare the generalization results of models with or without the continuous interpolation in Fig.~\ref{fig:continuous} (d).
% It can be seen that the continuous interpolation consistently improves the generalization performance on both OD/OC and prostate segmentations, owing to the benefits of learning domain-invariant updates in a more continuous distribution space.
We then analyze the effect of the choice of $\lambda$ on our model performance, for which we conduct experiments with fixed values from 0.0 to 1.0 with a step size 0.2, and continuous sampling in range of [0.0, 0.5], [0.5, 1.0] and [0.0, 1.0]. As shown in Fig.~\ref{fig:continuous} (b), compared with not transferring any distribution information (i.e., $\lambda=0$), setting $\lambda>0$ as a fixed value can always improve the model performance. Besides, the continuous sampling can further improve the performance and the sampling range of [0.0, 1.0] yields the best results, which reflects the benefits of continuous distribution space for domain generalization.%how the value of interpolation ratio $\lambda$ in Eq.~\ref{mixup} affects the generalization performance (cf. Fig.~\ref{fig:continuous} (b)).
% To compare with our strategy of dynamically generating $\lambda$ between [0.0, 1.0], we conduct experiments with a fixed value of $\lambda$ from 0.0 to 1.0 with a step size 0.2.
% With $\lambda=0.0$, we obtain the original data with no new distribution information of other domains.
% With $\lambda$ increasing from 0.0 to 1.0, the simulated data attain more and more distribution information from another domain, giving rise to higher generalization performance.
% Our proposed random sampling strategy achieves the best results, demonstrating that the continuous distribution space can better facilitate the model towards domain invariance than distinct distributions.

% With $\lambda=0.0$, we obtain the original data with no new distribution information of other domains, resulting in unsatisfactory generalization performance.
% It is observed that the generalization performance improves with the increase in $\lambda$ from 0.0 to 1.0, as the generated data attain more and more distribution information from other domains.
% Our proposed random sampling strategy for $\lambda$ achieves better results than the fixed value, demonstrating that better domain-invariance can be obtained in a more continuous distribution space than the distinct distributions.

% \begin{table}
% \caption{Performance of our method without continuous interpolation or without boundary-oriented meta regularization.}
% \label{tab:comparedetail}
% \resizebox{0.7\textwidth}{!}{%

%  \begin{tabular}{c|c c c}
%  \hline
%  Methods &Optic disc &Optic cup & Prostate  \\
%  \hline
%  \hline
%  Ours &92.69 &81.37 &87.39 \\
%  \hline
%  w/o continuous &92.02 &80.89 &86.87 \\
%  w/o boundary &91.89 &80.76 &86.53 \\
%  \hline
%  \end{tabular}
% }
% \end{table}

\begin{figure}[t]
	\centering
	\includegraphics[width=0.98\textwidth]{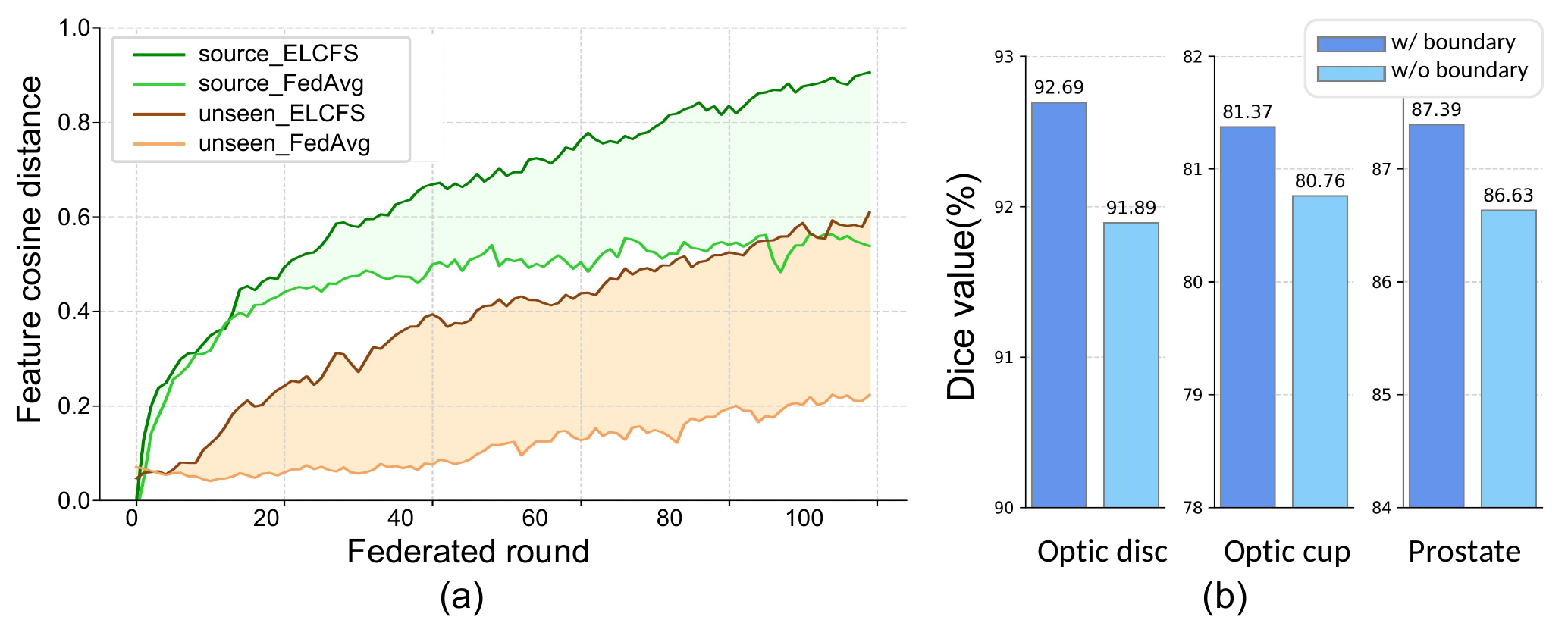}
	\caption{(a) Cosine distance between the boundary-related and background-related features; (b) Generalization performance of our method with or without the boundary-oriented meta objective.} 
	\label{fig:boundary_ablation}
	 		 \vspace{-1.5mm}
\end{figure}

\textbf{Discriminability at ambiguous boundary region:}
We plot the cosine distance between the boundary-related and background-related features, i.e., $\mathbb{E}[h_{i\_bd}\odot {h_{i\_bg}}]$, 
to analyze how the semantic feature space around the boundary region is influenced by our method.
%We plot the consine distances between the boundary-related and background-related features (cf.~\ref{eq:embeddings}), i.e., $\mathbb{E}[h_{l}\odot h_{n}-h_{l} \odot h_{p}]$, to analyze how the semantic feature space of the ambiguous region is influenced by our method. 
In Fig.~\ref{fig:boundary_ablation} (a), the two green lines denote the growth of feature distance in our ELCFS and the FedAvg baseline respectively, for samples drawn from the training source domains. 
We can see that ELCFS yields a higher feature distance, indicating that the features of the boundary and the surrounding background region can be better separated in our method.
For the two yellow lines, sample features are drawn from the unseen domains.
As expected, the distance  is not as high as in source domain, yet our method also presents a clearly higher margin than FedAvg.
We also quantitatively analyze the effect of $\mathcal{L}_{boundary}$ on the model performance.
As observed from Fig.~\ref{fig:boundary_ablation} (b), removing this objective from the meta optimization leads to consistent performance drops on the generalization performance in different tasks. 
% It is noted that the margin gain with our method is more consistently and obviously for samples from the unseen target domains than the training source domains, indicating that our method specifically works on improving the model's generalizability in unseen domains.

\begin{figure}[t]
	\centering
	\includegraphics[width=0.98\textwidth]{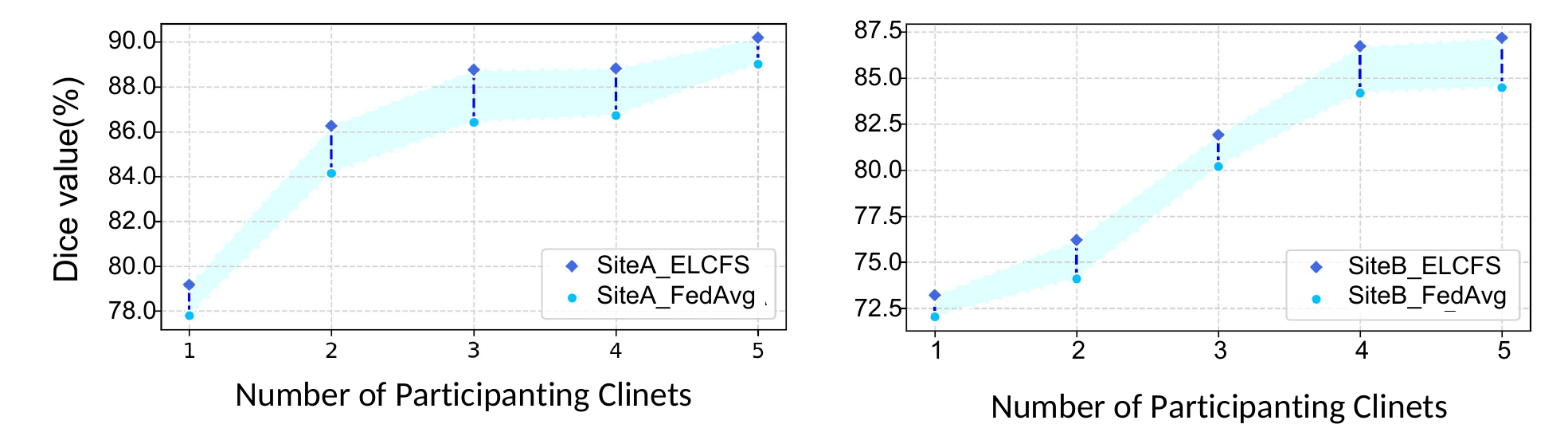}
	\caption{Curves of generalization performance on two unseen prostate datasets (i.e., site A and B) as the number of participating clients increases, using our proposed approach and FedAvg.} 
	\label{fig:client_number}
	 		 \vspace{-1.5mm}
\end{figure}

\textbf{Effect of participating client number:}
We further analyze how the generalization performance of our method and FedAvg will be affected when different numbers of hospitals participating in federated learning.
Fig.~\ref{fig:client_number} shows the results on prostate MRI segmentation, in which we present the generalization results on two unseen sites with the client number gradually increasing from 1 to $K-1$. 
As expected, the models trained with single-source data cannot obtain good results when deployed to unseen domains. 
The generalization performance increases when more clients participating in the federated training, which is reasonable as aggregating data from multiple sources can cover a more comprehensive data distribution.
Particularly, our ELCFS consistently outperforms FedAvg on all generalization settings with different client numbers, demonstrating the stable efficacy of our method to leverage distributed data sources to enhance the generalizability of federated learning model.

\section{Conclusion}
% \vspace{-1mm}
We have proposed a novel problem setting of federated domain generalization, and presented a novel approach for it with continuous frequency space interpolation and a boundary-oriented episodic learning scheme. The superior efficacy of our method is demonstrated on two important medical image segmentation tasks. Our solution has opened a door in federated learning to enable local client access multi-source distributions without privacy leakage, which has great potential to address other problems encountered in FL, e.g., data heterogeneity. The proposed learning scheme for encouraging boundary delineation is also generally extendable to other segmentation problems.

\section{Acknowledgement}
This work was supported by Key-Area Research and Development Program of Guangdong Province, China (2020B010165004); National Natural Science Foundation of China with Project No. U1813204; Hong Kong Innovation and Technology Fund (Project No. ITS/311/18FP and GHP/110/19SZ).

\clearpage
{\small
\bibliographystyle{ieee_fullname}

}

\clearpage

% \newpage
\begin{figure*}[t]
	\centering
	\includegraphics[width=\textwidth]{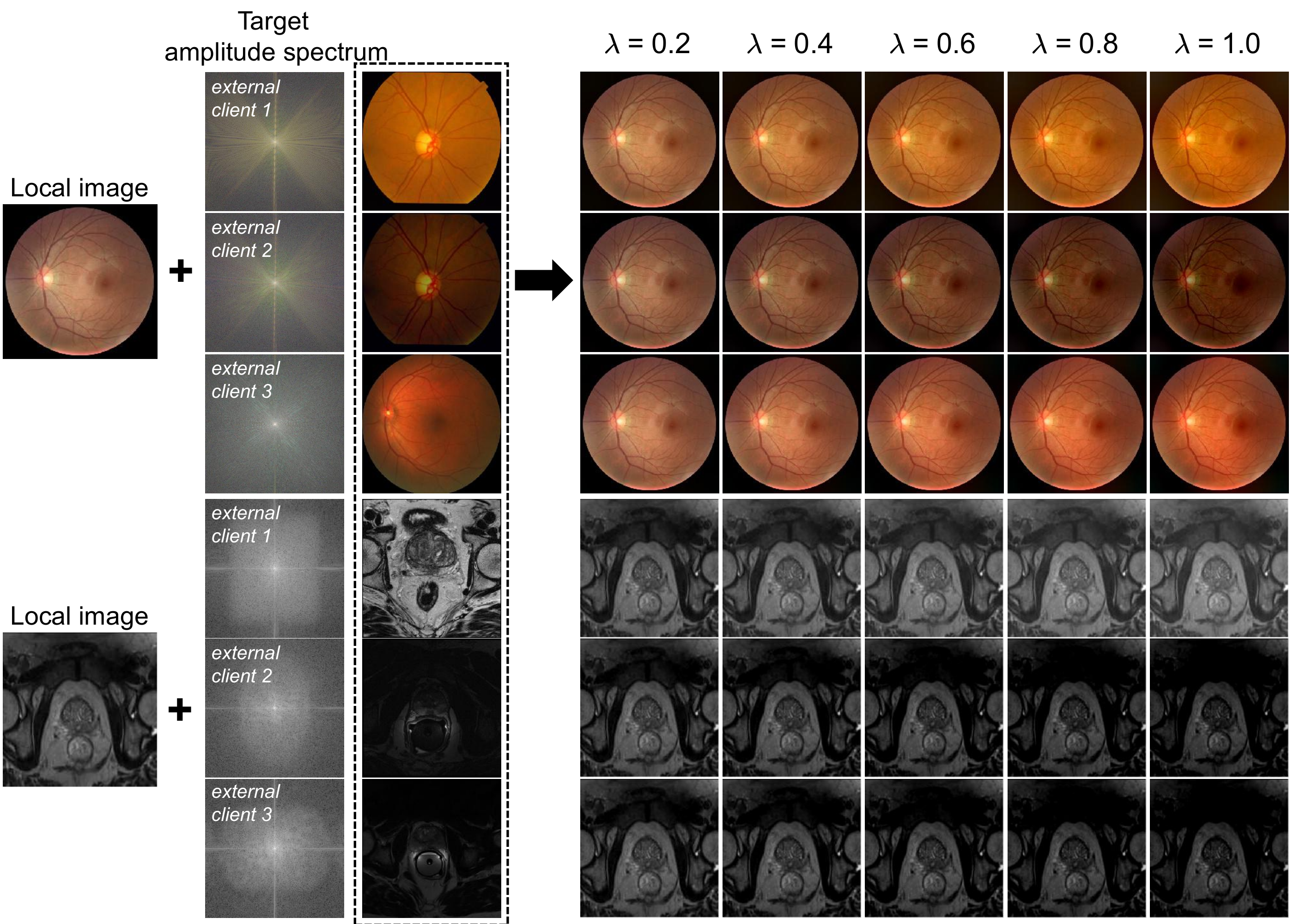}
	\caption{Visualization of transformed images under different interpolation ratio $\lambda$.}
% 	\caption{Overview of our proposed method for federated domain generalization on medical image segmentation.
% 	The distribution information is exchanged across clients from frequency space with an interpolation mechanism, enabling local client access the multi-source distributions in a continuous space. An episodic training paradigm is then established to expose the local optimization to transferred domain shift, with explicit regularization to promote  domain-independent feature cohesion and separation in the ambiguous boundary region.}
	\label{fig:ration}
\end{figure*}

\section{Supplementary Material}
\subsection{Datasets Details:}
The retinal fundus images adopted in the experiments are collected from four different clinical centers out of three public datasets. Among these data, samples of sites A are from Drishti-GS~\cite{sivaswamy2015comprehensive} dataset; samples of site B are from RIM-ONE-r3~\cite{fumero2011rim} dataset; samples of site C, D are from REFUGE~\cite{orlando2020refuge} dataset. Note that the REFUGE dataset includes two different data sources, so we decompose them in our federated learning setting. Among the six data sources in the prostate MRI segmentation task, samples of Site A, B are from NIC-ISBI13~\cite{bloch2015nci} datasets; samples of Site C are from I2CVB~\cite{lemaitre2015computer} datasets; and samples of Site D, E, F are from PROMISE12~\cite{litjens2014evaluation} dataset. Similarly, since the NIC-ISBI13 and PROMISE12 contain data from multiple data sources, we separate them and consider each data source as an individual client in the federated scenario. Details of the scanners and imaging protocols of these data are illustrated in Table~\ref{tab:dataset_fundus} and Table~\ref{tab:dataset_prostate} respectively.

\begin{table}[!h]
    \renewcommand\arraystretch{1.1}
    \centering
    \caption{\small{Details of the scanning protocols for different data sources in fundus image segmentation.}}
    \scalebox{0.7}{
    \begin{tabular}{c l  l l l l l }
    \hline
    Task &Dataset      & Manufactor \\
    % \hline
    \hline
    \multirow{4}{*}{\shortstack{Fundus \\ Image \\ Segmentation}} & Site A~\cite{sivaswamy2015comprehensive}   & (Aravind eye hospital)  \\
    & Site B~\cite{fumero2011rim}   & Nidek AFC-210   \\
    & Site C~\cite{orlando2020refuge}    & Zeiss Visucam 500\\
    & Site D~\cite{orlando2020refuge}    & Canon CR-2  \\
    \hline
    \end{tabular}
    }
\label{tab:dataset_fundus}
\end{table}
\begin{table}[!h]
    \renewcommand\arraystretch{1.1}
    \centering
    \caption{\small{Details of the scanning protocols for different data sources in prostate MRI segmentation.}}
    \scalebox{0.65}{
    \begin{tabular}{c l  l l l l l }
    \hline
    Task &Dataset  & Manufactor &Field strength(T)  & Endorectal Coil    \\
    \hline
    \multirow{6}{*}{\shortstack{Prostate \\ MRI \\ Segmentation}} & Site A~\cite{bloch2015nci}  & Siemens  & 3  &Surface\\
    & Site B~\cite{bloch2015nci}  & Philips   & 1.5 &Endorectal\\
    & Site C~\cite{lemaitre2015computer}  & Siemens  & 3 &No \\
    & Site D~\cite{litjens2014evaluation} & Siemens  & 1.5 and 3 &No\\
    & Site E~\cite{litjens2014evaluation} & GE  & 3 &Endorectal\\
    & Site F~\cite{litjens2014evaluation} & Siemens & 1.5 &Endorectal\\
    \hline
    \end{tabular}
    }
\label{tab:dataset_prostate}
\end{table}

\subsection{Statistical Analysis}
% We calculate the standard division (std) for the generalization results of different comparison methods. The results of  the two tasks are shown in Table~\ref{tab:comparison_fundus} and Table~\ref{tab:comparisons_prostate} respectively. We notice that in fundus image segmentation (cf. Table~\ref{tab:comparison_fundus}), the std with considering site A and site B as unseen sites are relatively higher than the others. The reason could be that generalizing to these two sites when training with remaining three sites are more difficult, causing that the generalization results present a larger cross-subject variance. For prostate MRI segmentation (cf. Table~\ref{tab:comparisons_prostate}), the std are relatively stable across different generalization settings compared with the the fundus image segmentation task.
We conduct paired t-test between our approach and different comparison methods to analyze whether the performance improvement of our approach is significant. We adopt Dice as the evaluation measurement and set the significance level as 0.05. 
% For each method, we concatenate its predictions under different unseen site settings to perform significance analysis on overall generalization performance. 
For each method, the statistical tests are conducted by jointly considering the prediction results of each unseen site setting on overall generalization performance.
The results are listed in Table~\ref{table:pair}. It is observed that all paired t-test results present \textit{p}-value smaller than 0.05, demonstrating that our improvements over these state-of-the-art domain generalization methods are significant.

\begin{table}[t]
    \renewcommand\arraystretch{1.2}
    \centering
        \caption{\small{P-value for statistical analysis between our approach and different comparison methods on overall Dice score.}}
        \label{tab:comparisons}
        \scalebox{0.7}{
        \begin{tabular}{c|c  c  c c c}
            \hline
              &JiGen~\cite{carlucci2019domain} & BigAug~\cite{zhang2020generalizing}  &Epi-FCR~\cite{li2019episodic} &RSC~\cite{huang2020self} &FedAvg~\cite{mcmahan2017communication}\\
            \hline
            \multirow{1}{*}{Optic disc} &3.6e-20 &7.6e-22  &3.5e-12 &2.1e-9 &1.2e-8\\
            \hline
            \multirow{1}{*}{Optic cup} &1.1e-16 &0.0026 &1.6e-7 &0.0003 &2.0e-21 \\
            \hline
            \multirow{1}{*}{Prostate} &0.0004 &2.3e-7 &9.2e-5 &5.2e-8 &2.9e-8 \\
          \hline
        \end{tabular}
    }
\label{table:pair}
\end{table}
\subsection{Visualization of Transformed Data}
We visualize the appearances of transformed images under different interpolation ratio $\lambda$ for the two tasks. As shown in Fig.~\ref{fig:ration}, the appearance of local source image is indeed  gradually transformed to the style (i.e. distribution) of target image of other clients as we increase the interpolation ratio from 0 to 1, while the semantic content of the image is unchanged. Such continuous interpolation mechanism helps to enrich the multi-source distributions to a dedicated dense distribution space, hence benefits the model to gain domain-invariance in a more continuous latent space to improve the generalizability. 

% \includepdf[pages=-]{Appendix.pdf}
% \clearpage
% \input{cvpr_supplementary}
\end{document}